\documentclass[journal,twoside,web]{ieeecolor}
\usepackage{generic}
\usepackage{cite}
\usepackage{amsmath,amssymb,amsfonts}
\usepackage{algorithmic}
\usepackage{graphicx}
\usepackage{textcomp}

\usepackage{epsfig}
\usepackage{epstopdf} 
\usepackage{multirow} 
\usepackage{graphicx}
\usepackage{wrapfig}
\usepackage{amsmath}
\usepackage{colortbl}
\usepackage{float}  
\usepackage{caption}
\usepackage{lipsum}
\usepackage{makecell}
\usepackage{color,xcolor}
\usepackage{booktabs}

\begin{document}
\title{FastTrackTr: Real-time Multi-Object Tracking with Transformers for Real World}
\author{Pan Liao,Feng Yang,Di Wu,Jinwen Yu,Wenhui Zhao,Dingwen Zhang}

\maketitle

\begin{abstract}

 Transformer-based multi-object tracking (MOT) methods have attracted significant attention from researchers. However, these Transformer-based models often suffer from suboptimal inference speeds due to their architectural complexities or other inherent issues, rendering them difficult to deploy in practical industrial applications. To address this challenge, we revisited the classic joint detection and tracking (JDT) paradigm and analyzed existing models. Drawing inspiration from DETR's object queries, which naturally encode object appearance features, we constructed a fast and novel JDT-type MOT framework named FastTrackTr by implementing an efficient inter-frame information transfer mechanism. Benefiting from the superiority of this mechanism, our approach not only reduces the number of queries required for tracking but also avoids introducing excessive network structures, ensuring model simplicity. Experimental results demonstrate that our method not only achieves competitive tracking accuracy across multiple datasets but also enables real-time tracking in practical application environments.

\end{abstract}


\section{Introduction}


\IEEEPARstart{M}{ulti}-object tracking is a critical research field in computer vision and industrial intelligence \cite{wang2023blockchain,yang2019efficient,tian2025tracking,xu2024ermot,xu2024rethinking}, with broad applications in security monitoring, small unmanned ground vehicles, video object analysis, crowd flow detection, and other domains. In recent years, some tracking methodologies based on transformer \cite{vaswani2017attention}  have garnered significant attention. Meanwhile, these models \cite{sun2020transtrack,meinhardt2022trackformer,zeng2022motr,yanBridgingGapEndtoend,gao2023memotr,yu2023motrv3,gao2024multiple} have also achieved state-of-the-art (SOTA) results on many benchmarks. However, these methods still suffer from a critical issue: their inference speeds remain suboptimal, and the network architectures of some models are highly inconvenient for engineering deployment, making them difficult to apply in practical industrial production. In contrast, some other existing methods \cite{maggiolinoDeepOCSORTMultiPedestrian2023,yang2024hybrid,cao2023observation} exhibit accuracy that may lag behind Transformer-based approaches. Therefore, developing Transformer architectures for real-time tracking remains an urgent research goal.

\begin{figure*}
	\centering
	\begin{minipage}[t]{0.55\linewidth}
				\centering
				\includegraphics[width=\linewidth]{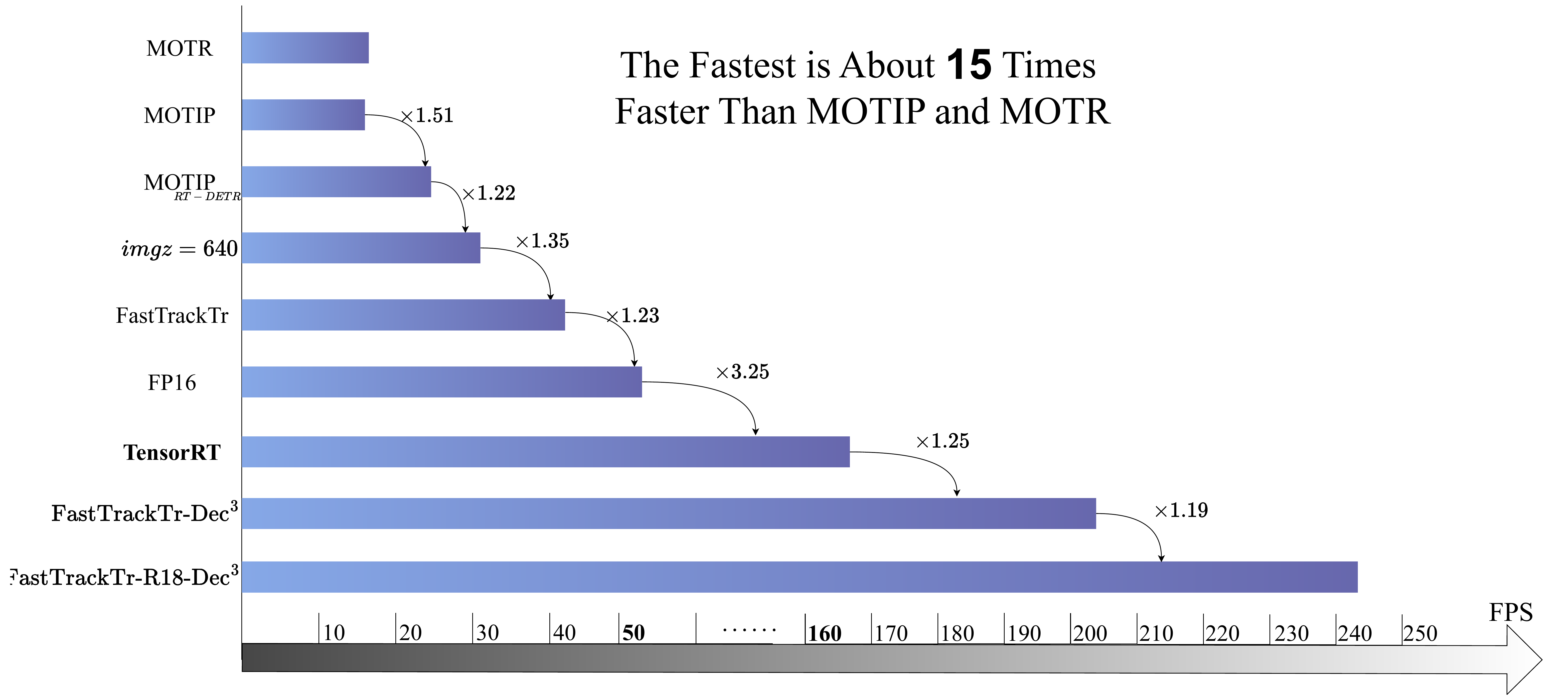}  
			\end{minipage}
	\hfill
	\begin{minipage}[t]{0.40\linewidth}
				\centering
				\includegraphics[width=\linewidth]{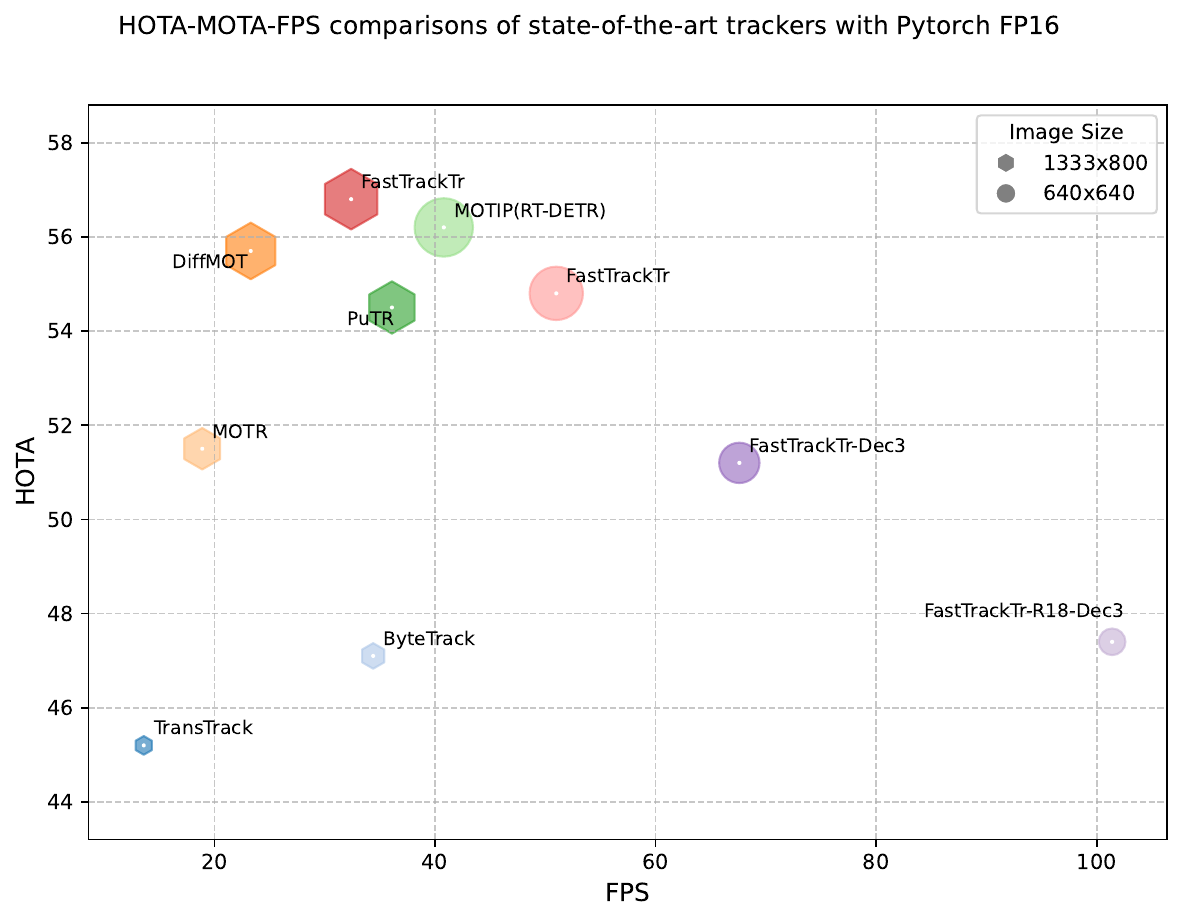}  
			\end{minipage}
	\caption{A comparison of the frames per second (FPS) and HOTA of various versions of FastTrackTr and the most advanced transformer-based models on a single NVIDIA RTX 4090. $\text{Dec}^3$ refers to a 3-layer decoder, while R18 refers to ResNet18. Additionally, RT-DETR was utilized to construct the MOTIP. The data precision of the scatter plot on the right is PyTorch fp16 and the area of points represents MOTA.}
	\label{fps}
\end{figure*}

 The limited inference speed of existing methods stems from multiple factors. For query-based tracking models like the MOTR series \cite{zeng2022motr,yanBridgingGapEndtoend,gao2023memotr,yu2023motrv3}, the primary bottleneck lies not in network architecture design but in hardware acceleration challenges caused by variable query quantities. Notably, computational complexity in many modules scales quadratically with the number of queries. Both our empirical studies and other work \cite{pan2023mo} confirm that network lightweighting alone yields insufficient speed improvements. The more recent method, MOTIP \cite{gao2024multiple} addresses several MOTR limitations but introduces an auxiliary ID prediction network. Even when substituting MOTIP's Deformable DETR \cite{zhu2020deformable} with RT-DETR \cite{zhao2024detrs}, inference latency remains unsatisfactory (results shown in Fig. \ref{fps}). Furthermore, MOTIP's dynamic parameterization complicates acceleration through TensorRT optimization. Tracking-by-Detection (TBD) paradigms \cite{maggiolinoDeepOCSORTMultiPedestrian2023,yang2024hybrid}, while computationally efficient and deployment-friendly, face their own constraints. When accounting for the combined latency of detection models and re-identification (ReID) networks, only purely kinematic methods \cite{cao2023observation} achieve competitive speeds—at the cost of inferior performance compared to ReID-enhanced approaches. This raises a fundamental question: Can we develop a framework that simultaneously achieves speed, precision, and practical deployability?

Revisiting historical solutions provides critical insights. The Joint Detection and Tracking (JDT) paradigm was originally designed for rapid tracking but has fallen out of favor in recent years. Nevertheless, its core principles may hold the key to resolving current challenges. Existing transformer-based JDT implementations like TransTrack \cite{sun2020transtrack} suffer from limitations similar to MOTIP, while other query-based methods \cite{meinhardt2022trackformer,xu2022transcenter,cai2022memot} inherit MOTR's fundamental constraints. CNN-based JDT methods \cite{zhang2021fairmot,wang2020towards}, though now less prevalent due to outdated base architectures, demonstrated an effective design: simultaneous detection output and ID embedding generation, followed by association matching. This paradigm inherently preserves inference efficiency while enabling deep coupling of detection features and identity representations through end-to-end joint training. Crucially, JDT aligns naturally with DETR architectures—the object queries produced by DETR inherently encode spatial-semantic information, thereby circumventing the tracking-by-query limitations when repurposed as ID embeddings. Moreover, transformer attention mechanisms surpass CNN counterparts in processing implicit temporal correlations across video frames, potentially enhancing tracking accuracy.

Building upon these insights, Based on the above discussion, we present \textbf{FastTrackTr}, a \textbf{Fast} multi-object \textbf{Track}ing method grounded in the \textbf{Tr}ansformer architecture and aligned with the JDT paradigm. This tracking method achieves impressive tracking progress through a unique decoder that integrates past information. Additionally, due to the ease of generating masks for this method, the batch size can be set relatively large, making our model considerably more training-friendly compared to some of the latest transformer-based MOT methods. Furthermore, the deterministic data architecture inherent in FastTrackTr enables seamless TensorRT acceleration, achieving genuine real-time performance. This computational efficiency remains largely unattainable for contemporary transformer-based MOT methods due to their inherent dynamic computation patterns.

In summary, the main innovations of this paper are as follows:
\begin{itemize}
	\item [1)] proposing \textbf{FastTrackTr}, a novel transformer-based MOT method that achieves real-time inference speed while maintaining high accuracy.
	\item [2)] introducing a cross-decoder mechanism that implicitly integrates historical trajectory information, eliminating the need for additional queries or decoders.
	
	\item [3)] conducting extensive experiments on multiple benchmark datasets, including DanceTrack \cite{sunDanceTrackMultiObjectTracking2022}, SportsMOT \cite{cui2023sportsmot}, MOT17 \cite{milanMOT16BenchmarkMultiObject2016}, Visdrone2019 \cite{zhu2021detection}, and BDD100k \cite{yu2020bdd100k}, to demonstrate its robust adaptability to various environmental conditions. 
    
    \item [4)] performing physical experiments in industrial settings to confirm its practical feasibility for real manufacturing applications.
	
\end{itemize}

\section{Related work}


\subsection{Transformer-based MOT Methods}
The earliest Transformer-based MOT methods were TransTrack \cite{sun2020transtrack} and TrackFormer \cite{meinhardt2022trackformer}. The former is a JDT model, while the latter introduced the concept of track queries, utilizing these special queries to achieve tracking. Subsequently, MOTR \cite{zeng2022motr} proposed tracklet-aware label assignment (TALA), which optimized label assignment during the training phase to achieve end-to-end tracking. Later improvements to MOTR, such as MOTRv2 \cite{zhang2023motrv2}, MeMOTR \cite{gao2023memotr}, MOTRv3 \cite{yu2023motrv3}, and CO-MOT \cite{yanBridgingGapEndtoend}, demonstrated the powerful tracking performance of the MOTR paradigm. Some efforts have been made to address MOTR's slow inference speed through model lightweighting, but the results have been less than ideal.
Recently, alternative Transformer architectures have been proposed. MOTIP's \cite{gao2024multiple} novel approach of directly predicting tracked object ID numbers is refreshing, while PuTR's \cite{liu2024putr} use of Transformers to directly construct a tracking-by-detection paradigm tracker has also been inspirational. 

\subsection{Tracking-by-Detection}

The TBD paradigm can be considered one of the simplest and most practical tracking methods. Since it essentially associates detected targets, this class of methods often yields satisfactory tracking accuracy as long as the detection results are not poor. The initial SORT algorithm simply utilized the motion state of detected objects for association. Subsequent models like ByteTrack \cite{zhangByteTrackMultiobjectTracking2022} improved tracking accuracy by enhancing detection precision and employing embedding-based Re-identification (Re-id) networks. Some approaches, like OC-SORT \cite{cao2023observation} and HybridSORT \cite{yang2024hybrid}, improved upon SORT using purely kinematics-based methods. Later models such as DeepOC-SORT \cite{maggiolinoDeepOCSORTMultiPedestrian2023} demonstrated that simultaneous improvements in both Re-id network utilization and kinematic models yield the most significant performance enhancements. Currently, there are also algorithms that directly use networks for association. 

\subsection{Joint Detection and Tracking}

Joint detection and tracking aims to achieve detection and tracking simultaneously in a single stage. This approach was common a few years ago but has been seen less frequently in recent years. Here, we briefly review its development history. JDE \cite{wang2020towards} and FairMOT \cite{zhang2021fairmot} learned object detection and appearance embedding tasks from a shared backbone network. CenterTrack \cite{zhou2020tracking} localized targets through tracking-conditioned detection and predicted their offsets from the previous frame. MMTrack\cite{xu2024rethinking} has constructed a JDT model suitable for multiple scenarios. TransTrack used two decoders to handle tracking and detection separately, then matched and associated the results from both decoders. Our FastTrackTr, by introducing a novel decoder, returns to the essence of the JDT method, simplifies the network structure, and achieves fast tracking.

\section{Method}

\subsection{FastTrackTr Architecture}

\begin{figure*}[t]
	\begin{center}
		\includegraphics[width=0.8\linewidth]{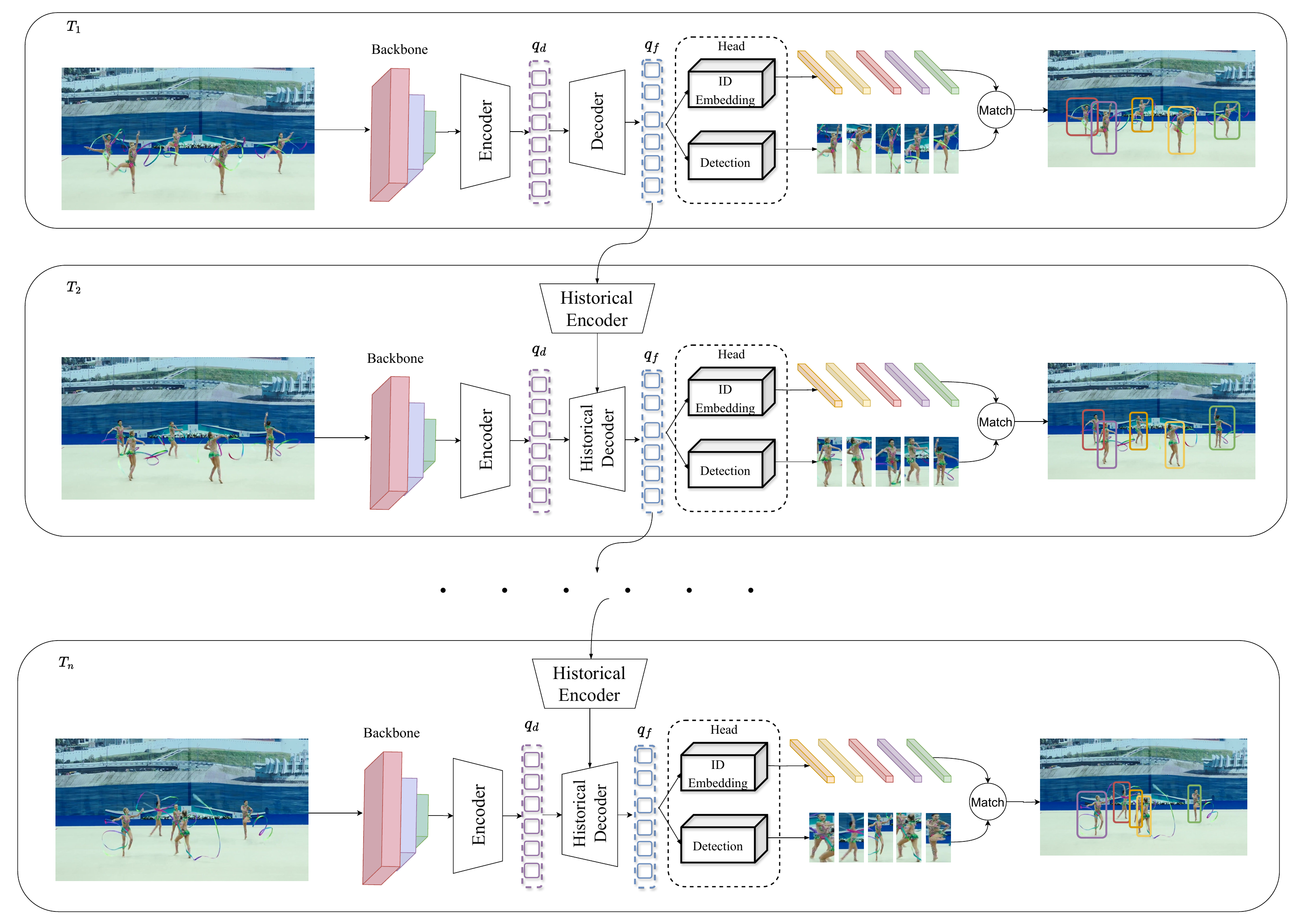}
	\end{center}
	\caption{The complete structure of FastTrackTr. For the initial frame, the model functions as a detection system, assigning unique identifiers to each detected object. Subsequently, from the second frame onwards, the historical queries from previous frames and their corresponding masks undergo processing through a single-layer encoder. }
	\label{Architecture}
\end{figure*}

The overall structure of FastTrackTr is illustrated in Fig. \ref{Architecture}. In frame $T_1$, a standard decoder is used to initialize the initial historical information. Subsequently, in the following frames, a Historical Encoder is employed to initialize this historical-representative information. In the next frame, a Historical Decoder, which includes a historical cross-attention layer compared to normal decoder layers, processes this historical information. The tracking principle is very similar to that of JDE and FairMOT, utilizing an additional ID Embedding head to obtain the object's appearance embedding, which is then used along with detection results for association and matching. This architecture achieves efficient tracking, which can be run in real time on edge devices.



\subsection{Learning Detection and ID Embedding}
%

To enable our FastTrackTr to simultaneously output both the target's position and id embedding in a single forward pass, ensuring optimal inference speed, we formulate the problem as follows: Given a training dataset $\{(\mathbf{I}_i, \mathbf{B}_i, \mathbf{y}_i)\}_{i=1}^N$ consisting of image frames $\mathbf{I}_i \in \mathbb{R}^{C\times H\times W}$, bounding box annotations $\mathbf{B}_i \in \mathbb{R}^{K_i\times 4}$, and partial identity labels $\mathbf{y}_i \in \mathbb{Z}^{K_i}$, we aim to develop a unified model that simultaneously predicts bounding boxes $\tilde{\mathbf{B}}_i \in \mathbb{R}^{\tilde{K}_i\times 4}$ and discriminative appearance embeddings $\tilde{\mathbf{F}}_i \in \mathbb{R}^{\tilde{K}_i\times D}$ through a single forward pass.

The model must satisfy two fundamental requirements: (1) accurate bounding box regression through minimization of the localization loss $\mathcal{L}_{\text{det}}(\tilde{\mathbf{B}}, \mathbf{B},\tilde{\mathbf{y}}, \mathbf{y})$, and (2) temporally consistent feature embeddings where the distance $d_\phi(f_t^k, f_{t+\Delta t}^m)$ between matching identities ($\mathbf{y}_t^k = \mathbf{y}_{t+\Delta t}^m$) is smaller than non-matching pairs ($\mathbf{y}_t^k \neq \mathbf{y}_{t+\Delta t}^{m'}$) by a margin $\alpha > 0$. Formally, the joint objective can be expressed as:

\begin{equation}
	\min_{\theta,\phi} \mathbb{E}\left[\mathcal{L}_{\text{det}} + \lambda \max(0, d_\phi(f_t^k, f_{t+\Delta t}^m) - d_\phi(f_t^k, f_{t+\Delta t}^{m'}))\right]
\end{equation}

where $\lambda$ controls the balance between detection accuracy and feature discriminability, with $f_t^k$ denoting the $k$-th row vector of $\tilde{\mathbf{F}}_t$ corresponding to the $k$-th predicted instance.

Our framework builds upon the DETR architecture \cite{zhu2020deformable}, selected for its inherent compatibility with our technical objectives. The Transformer decoder's parallel decoding mechanism transforms learnable object queries into interpretable target representations through cross-attention interactions with image features. These joint representations intrinsically encode spatial coordinates and semantic features, which are conventionally mapped to classification probabilities and bounding box coordinates via separate prediction heads.

We extend this paradigm through two strategic modifications: (1) Introducing an auxiliary \textit{ID embedding head} implemented as a fully-connected layer that projects object queries into a discriminative feature space for target association, and (2) Implementing a joint optimization strategy that preserves detection performance while learning identity-aware representations. The base detection component employs RT-DETR \cite{zhao2024detrs} for its real-time efficiency, maintaining its original dual-branch structure and loss formulation:

\begin{equation}
	\mathcal{L}_{\text{det}} = \underbrace{\mathcal{L}_{\text{box}}(\hat{b}, b)}_{\text{Localization}} + \underbrace{\mathcal{L}_{\text{cls}}(\hat{c}, c, \text{IoU})}_{\text{Classification}}
\end{equation}

where $\hat{b}, b$ denote predicted/ground-truth boxes and $\hat{c}, c$ represent class predictions. The classification loss integrates IoU-aware weighting following RT-DETR, while the box regression combines L1 and generalized IoU losses.

Previous JDT methods \cite{wang2020towards,zhang2021fairmot} suffer from scalability limitations due to their one-hot identity encoding schemes. We instead formulate identity embedding learning as a metric learning problem, decoupling feature dimensions from identity quantities to enhance cross-scene generalizability. This contrasts with recent motion-centric approaches \cite{gao2024multiple,liu2024putr} that underutilize spatial features encoded in DETR queries.

Our method employs Circle Loss \cite{sun2020circle} with adaptive margin control to optimize pairwise similarities in the embedding space. Specifically, by utilizing the \textbf{dot-product similarity}, the id embeddings of the same object across multiple frames are regarded as positive pairs indicated as $s_p$, while those between distinct objects are treated as negative samples represented by $s_n$. - $\alpha_p$ and $\alpha_n$ are the adaptive weighting factors for positive and negative samples, determined by the following equations:

\begin{equation}
	\begin{aligned}
		\alpha_p &= \max(0, O_p - s_p), \quad \alpha_n = \max(0, s_n - O_n) \\
		O_p &= 1 + m, \quad O_n = -m \quad (m > 0)
	\end{aligned}
\end{equation}

where $m$ controls the separation margin. The final loss formulation:

\begin{equation}
	\mathcal{L}_{\text{reid}} = \log\left[1 + \sum_j e^{\gamma\alpha_n(s_n^j - \Delta_n)} \sum_i e^{-\gamma\alpha_p(s_p^i - \Delta_p)}\right]
\end{equation}

with $\Delta_p = 1 - m$, $\Delta_n = m$, and $\gamma$ as the scaling factor. This adaptive weighting mechanism prioritizes hard samples in both positive and negative groups.

The complete objective function combines detection and metric learning losses through layer-wise aggregation across decoder stages:

\begin{equation}
	\mathcal{L}_{\text{total}} = \lambda_{\text{det}}\mathcal{L}_{\text{det}} + \lambda_{\text{reid}}\mathcal{L}_{\text{reid}}
\end{equation}
where $\lambda_{\text{det}} = \lambda_{\text{reid}} = 1$ ensures balanced optimization. At the same time, in order to better enable the model to implicitly learn temporal information, like other Transformer-based MOT models \cite{zeng2022motr,gao2024multiple,meinhardt2022trackformer,sun2020transtrack}, the loss here is updated after accumulating several consecutive frames.

\subsection{Historical Decoder and  Encoder}
While preceding sections have detailed the construction of an efficient Transformer-based MOT framework, the inherent temporal aggregation capabilities of Transformers remain underexploited in our current implementation. This section introduces a systematic methodology for efficiently leveraging historical information within the current frame.

\begin{figure}[t]
	\begin{center}
		\includegraphics[width=1\linewidth]{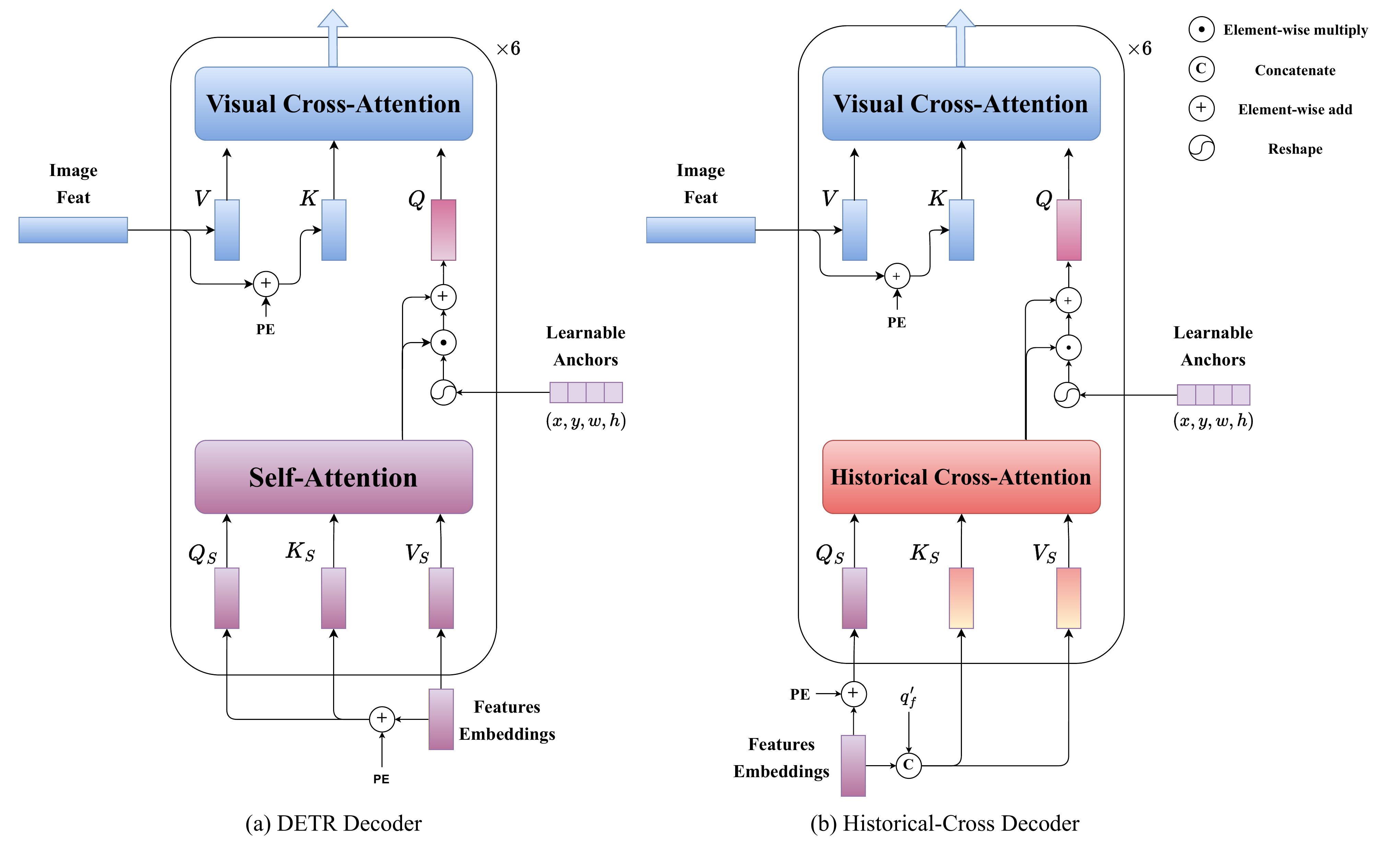}
	\end{center}
	\caption{Schematic of the flow of our historical-cross decoder and a general decoder. Compared to a general decoder, ours replaces self-attention with historical cross-attention. In our historical cross-attention mechanism, the setting of $q$ remains the same, while $k$ and $v$ incorporate information from historical frames.}
	\label{decoder}
\end{figure}

\textbf{Historical Decoder:} Previous DETR-based MOT models \cite{zeng2022motr,meinhardt2022trackformer,gao2023memotr} typically feed the final output queries of the previous frame, after some processing, directly into the decoder of the current frame to achieve tracking. This is an explicit tracking method, but it increases the number of queries, thereby increasing computational costs. To more efficiently utilize these historical queries, we replaced the self-attention mechanism in DETR's decoder with a cross-attention mechanism to process the query features between consecutive frames. This significantly reduces the computational load. The differences between our decoder and the standard decoder are shown in Fig. \ref{decoder}.

Specifically, each decoder module in DETR \cite{zhu2020deformable} consists of a sequential structure comprising a self-attention layer for the queries, a visual cross-attention layer, and a feed-forward neural network (FFN). Initially, the self-attention layer performs global modeling over the queries to enhance the model’s object detection performance. To further leverage historical information and improve tracking accuracy, we concatenate the Historical Encoder output from frame $t-1$, denoted as $q^{t-1’}_f$, which is processed by the trajectory encoder, with the initial queries $q_d^t$ of frame $t$. This concatenation replaces the key and value in the self-attention layer while keeping the queries unchanged. The implementation can be formally expressed as follows:

\begin{align}
	Q_T=\operatorname{Linear}(q_d^t),\quad K_T=V_T=\operatorname{Linear}(concat(q_d^t,q^{t-1’}_f))
\end{align} 
where $Q_T \in \mathbb{R}^{N \times C}$ , $K_T \text{ and } V_T \in \mathbb{R}^{2N \times C}$. Then, we calculate the query-trajectory attention map $A_T \in \mathbb{R}^{2N \times N}$, and aggregate informative depth features weighted by $A_T$ to produce the Historical-aware queries $q^{\prime}$, formulated as:

\begin{align}
	\begin{aligned}
		A_T & =\operatorname{Softmax}\left(Q_T K_T / \sqrt{C}\right) \\
		q^{\prime} & =\operatorname{Linear}\left(A_T V_T\right)
	\end{aligned}
\end{align} 
This mechanism allows each object query to adaptively capture features of the same object from images across consecutive frames, thereby achieving better object understanding. The queries with estimated historical information are then input into the inter-query self-attention layer for feature interaction between objects, and the visual cross-attention layer for collecting visual semantics. 

\textbf{Historical Encoder:}
In FastTrackTr, we have introduced an additional Historical Encoder to enhance the modeling of temporal relationships, and we utilize a masking mechanism during the self-attention process to provide contextual priors for the tracked objects. For the historical feature information from the previous frame or multiple frames, only a subset of the data is utilized. To effectively model these historical data, we employ a masking mechanism and process the data through standard encoder layers. During training, the generation of masks is determined by the ground truth labels, while in inference, masks are based on confidence scores to decide which objects to retain.

\begin{figure}[t]
	\begin{center}
		\includegraphics[width=\linewidth]{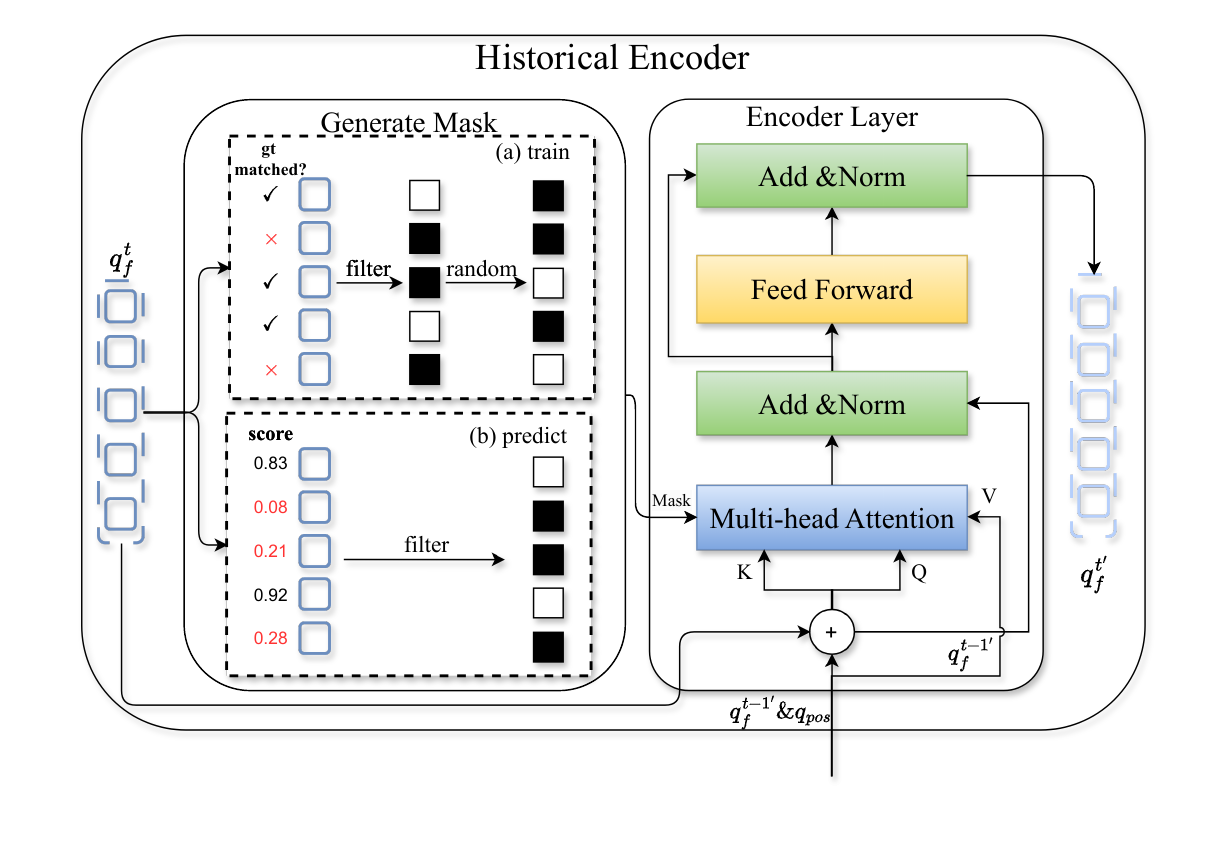}
	\end{center}

	\vspace{-0.3cm}
	\caption{Schematic of the flow of our Historical Encoder. The main role of this Encoder is to further aggregate the temporal features. The generated mask is randomly inverted partially during training. The probability of reversal is related to the number of truth values.  }
	\label{encoder_mask}
	\vspace{-0.3cm}
\end{figure}

The workflow of the historical encoder is illustrated in Fig. \ref{encoder_mask}. After generating the mask, we fuse the decoder output \(q_f^t\) with the historical information \(q_f^{t-1'}\) through an attention mechanism, computed as follows:
\begin{align}
	Q_E=K_E=\operatorname{Linear}(q^t_f+q_{pos}) , V_E=\operatorname{Linear}(q^{t-1'}_f)
\end{align}
where $q_{pos}$ represents positional encodings and the attention weights are computed via scaled dot-product as previously used in the decoder, followed by masking to suppress less relevant historical elements. Finally, the fused features are then processed by a feed-forward network to obtain \( q^{t'}_f \). For \( t=1 \), we initialize \( q^{t-1'}_f \) as \( q^t_f \) due to the absence of prior history.


To prioritize significant historical information, we design a dynamic masking strategy with two operational modes: (1) In the training phase, the mask is determined by ground truth information, generated by checking if each object matches with the ground truth labels, and random sampling of the masks is performed to flip them, enhancing the model's robustness; (2) In the inference phase, the mask is determined based on the confidence score of each object, with objects below a certain threshold being filtered out, while high-confidence objects are retained.  The mask adopts \cite{vaswani2017attention}'s softmax suppression via large negative biases to irrelevant entries. It strengthens the encoder's capability to model temporal dependencies while suppressing error propagation from noisy historical inputs.


\subsection{Matching and Association}
Although matching and association and not the primary focus of our paper, it is essential to outline some details of this process. Following the work of JDE, for a given video, after processing each frame, the model outputs detection results along with their corresponding ID embeddings. We then use these values to compute a similarity matrix between the embeddings of the observations and those of the existing tracking chains. The Hungarian algorithm is employed to assign observations to tracking chains. Subsequently, a Kalman filter is used to smooth the trajectories and predict the position of the previous frame's tracking chain in the current frame. If the assigned observations are spatially distant from the predicted position, the assignment is rejected. Finally, the embeddings of the tracking chains are updated as follows:

\begin{align}
	f_t=\eta f_{t-1}+(1-\eta) \tilde{f}
\end{align}
where $\tilde{f}$ indicates the embedding of the assigned observation, and $f_t$ indicates the embedding of the tracklet at frame $t$,  $\eta$ is a momentum term for smoothing, and we set $\eta=0.9$. 

Meanwhile, to further improve the model's performance, similar to MMTrack \cite{xu2024rethinking}, we adopted the iterative strategy of BYTE \cite{zhangByteTrackMultiobjectTracking2022} and ID features, which is consistent with the proposals of \cite{zhang2021fairmot,cao2023observation}.

\section{Experiments}

\begin{table*}
	\centering
	\caption{Overall Latency comparison between FastTrackTr and existing methods on the Dancetrack \cite{sunDanceTrackMultiObjectTracking2022} val. * denotes that these methods employ the same YOLOX \cite{ge2021yolox} detector, and the latency of this detector is accounted for in this table. Similarly, the latency of our method accounts for its associated time consumption, and these are highlighted in color. Unless otherwise specified, all Transformer-based models have ResNet50 for Backbone and 6 layers for decoder besides PuTR as it is a method of TBD paradigm. Not providing methods to accelerate TensorRT implies that they may not be well-suited for it.}
	\label{result_fps}
	\resizebox{\textwidth}{!}{
\begin{tabular}{l|c|c|c|c|c}
	\hline
	\textbf{Methods} & \textbf{Image Res.} & \textbf{Numerical Precision} & \textbf{Overall Latency(ms)} & \textbf{HOTA} & \makecell[c]{\textbf{Overall Latency with} \\ \textbf{TensorRT FP16(ms)}} \\
	\hline
	DiffMOT*\cite{lv2024diffmot} & 1333×800 & FP16 & 43.3ms (23.1fps) & 55.7 & 22.3ms(44.8fps) \\
	PuTR*\cite{liu2024putr} & 1333×800 & FP32/FP16 & 29.9ms (33.4fps) / 27.7ms (36.1fps) & 54.5 & 15.8ms(63.3fps) \\
	MOTR\cite{zeng2022motr} & 1333×800 & FP32/FP16 & 60.0ms (16.5fps) / 52.9ms (18.9fps) & 51.3 & - \\
	MOTIP\cite{gao2024multiple} & 1333×800 & FP32/FP16 & 61.7ms (16.2fps) / 45.7ms (21.8fps) & \textbf{59.1} & - \\
	
	\rowcolor{green!7} 
	FastTrackTr & 1333×800 & FP32/FP16 & 40.2ms (24.8fps) / 30.9ms (32.4fps) & 56.9 & 11.6ms(86.6fps) \\
	
	MOTIP(RT-DETR) & 640×640 & FP32/FP16 & 32.1ms (31.2fps) / 24.5ms (40.8fps) & 56.2 & - \\
	
	\rowcolor{green!7} 
	FastTrackTr & 640×640 & FP32/FP16 & 23.6ms (42.3fps) / 19.6ms (51.0fps) & 54.8 & 6.0ms(166.4fps) \\
	
	\rowcolor{green!14} 
	FastTrackTr-Dec\textsuperscript{3} & 640×640 & FP32/FP16 & 20.4ms (49.0fps) / 14.8ms (67.6fps) & 51.2 & 4.9ms(203.7fps) \\
	
	\rowcolor{green!20} 
	FastTrackTr-R18-Dec\textsuperscript{3} & 640×640 & FP32/FP16 & \textbf{12.8ms (78.0fps) / 9.86ms (101.4fps)} & 47.4 & \textbf{4.1ms(243.4fps)} \\
	\hline
\end{tabular}
		}
\end{table*}

\subsection{Setting}
\textbf{Datasets:}  To evaluate the tracking performance of FastTrackTr in real-world scenarios, we selected five datasets for testing: Dancetrack \cite{sunDanceTrackMultiObjectTracking2022}, SportMOT \cite{cui2023sportsmot}, MOT17 \cite{milanMOT16BenchmarkMultiObject2016}, BDD100k \cite{yu2020bdd100k}, and VisDrone2019 \cite{zhu2021detection}. These datasets comprehensively encompass most real-world scenarios—for instance, MOT17 for surveillance, BDD100k for driving, VisDrone2019 for drone-based applications, and DanceTrack/SportMOT for sports-related contexts. Finally, we also tested our model on our own physical platform.

\textbf{Metrics}: We assessed our method using widely recognized MOT evaluation metrics. The primary metrics employed for evaluation included HOTA, AssA, DetA ,IDF1 and MOTA \cite{ristani2016performance,luiten2021hota}. These metrics collectively provided a comprehensive and robust evaluation of our method's tracking performance. We also present additional evaluation metrics, including false positives (FP), false negatives (FN), identity switches (IDS), detection accuracy (DetA), association accuracy (AssA), and frame rate (FPS). And, for BDD100K, to better evaluate the performance of multiclass and multiobject tracking, we use tracking accuracy of all things (TETA) \cite{yu2020bdd100k}, localization accuracy (LocA), association accuracy (AssocA) and classification accuracy (ClsA) metrics.

\textbf{Implementation Details}: 
For inference speed, our FastTrackTr is built on RT-DETR \cite{zhao2024detrs}, with ResNet50 as the backbone unless specified otherwise. The model is implemented in PyTorch, and the training is conducted on 8 NVIDIA RTX 4090 GPUs. The sequence length for training is set to 8 unless otherwise noted. All ablation experiments use an image size of $640 \times 640$. For further testing of the best tracking accuracy of our method, we use an image size of $800 \times 1333$ when comparing with SOTA methods. The Historical Encoder during training applies random mask transformations, with the probability related to the number of ground-truth objects. Given the number of ground-truth objects $n_{gt}$ and the number of queries $n_q$, the probability of a non-masked flip is $ \min(0.5, 2 \cdot n_{gt} / n_q) $, while for the rest, the probability is $ \min(\max(0.1, n_{gt} / n_q), 0.2) $. For all three datasets, $n_q$ is set to 300. Dancetrack, BDD100k and SportMOT are trained for 27 epochs, while MOT17 and VisDrone2019 are trained for 80 epochs. The optimizer is AdwmW with a learning rate of $10^{-4}$ and weight decay of $5.0 \times 10^{-4}$. Due to GPU memory limitations, the batch size is set to 4 for smaller images and 1 for larger images on one GPU. Lastly, for all datasets, the final output dimension of ID embedding is 256. All models are implemented in PyTorch by default unless explicitly stated otherwise. The complete codebase remains exclusively Python throughout our experiments.

\subsection{Main Result}

\textbf{Comprehensive Comparison of Speed and Accuracy:} In RT-DETR, a standardized benchmark has been established for real-time detectors. Within this benchmark, all models undergo acceleration via TensorRT, with computational latency measurements exclusively encompassing inference and post-processing durations. This paper adopts the same configuration while substituting the detection dataset with the MOT dataset Dancetrack . However, this substitution reveals a critical implementation challenge: many transformer-based MOT methods \cite{sun2020transtrack,meinhardt2022trackformer,zeng2022motr,yanBridgingGapEndtoend,gao2023memotr,yu2023motrv3,gao2024multiple} developed in PyTorch incorporate dynamic computational graphs during inference, presumably to reduce computational overhead or simplify implementation. These dynamic elements create significant compatibility issues (specifically non-fixed tensor shapes and conditional branching) when converting models to TensorRT - challenges that remain unresolved in our current implementation. To ensure fair comparison, Table \ref{result_fps} comprehensively presents both speed and accuracy metrics across tracking methods using various numerical precisions in PyTorch. Notably, for FastTrackTr and the TBD method, only the network benefits from TensorRT acceleration, while the association module retains its original Python implementation and similar to real-time detection, in this paper, we only measure the time consumption of the tracking system, and the time for data preprocessing and post-processing is not included.

\begin{table}
	\centering
	\caption{Performance comparison between FastTrackTr and existing methods on the Dancetrack \cite{sunDanceTrackMultiObjectTracking2022} test set. The best performance among the methods is marked in \textbf{bold} and please pay more attention to the metrics with *.}
	\label{result_dancetrack}
	\resizebox{\columnwidth}{!}{%
		\begin{tabular}{l|*{5}{c}}
			\toprule
			\textbf{Methods} & \textbf{HOTA*} & \textbf{DetA} & \textbf{AssA} & \textbf{MOTA} & \textbf{IDF1} \\
			\midrule
			FairMOT\cite{zhang2021fairmot}        & 39.7    & 66.7     & 23.8     & 82.2     & 40.8 \\
			TransTrack\cite{sun2020transtrack}    & 45.5    & 75.9     & 27.5     & 88.4     & 45.2 \\
            MMTrack\cite{xu2024rethinking}      & 46.5    & 69.9      & 31.0     & 86.1 & 48.3 \\

			ByteTrack\cite{zhangByteTrackMultiobjectTracking2022} & 47.7    & 71.0     & 32.1     & 89.6     & 53.9 \\
			QDTrack\cite{fischer2023qdtrack}      & 54.2    & 80.1     & 36.8     & 87.7     & 50.4 \\
			MOTR\cite{zeng2022motr}               & 54.2    & 73.5     & 40.2     & 79.7     & 51.5 \\
			OC-SORT\cite{cao2023observation}      & 55.1    & 80.3     & 38.3     & 92.0     & 54.6 \\
			PuTR\cite{liu2024putr}                & 55.8    & $/$      & $/$      & 91.9     & 58.2 \\
			Deep OC-SORT\cite{maggiolinoDeepOCSORTMultiPedestrian2023}             & 61.3    & 82.2 & 45.8     & 92.3     & 61.5 \\
			DiffMOT\cite{lv2024diffmot}      & 62.3    & \textbf{82.2}      & 47.2     & \textbf{92.8} & 63.0 \\
			\rowcolor{green!8}\textbf{FastTrackTr(ours)}            & \textbf{62.4} & 78.4     & \textbf{49.8} & 88.8     & \textbf{64.8} \\
			\bottomrule
		\end{tabular}
	}
\end{table}

\begin{figure*}[t]
	\begin{center}
		\includegraphics[width=1\linewidth]{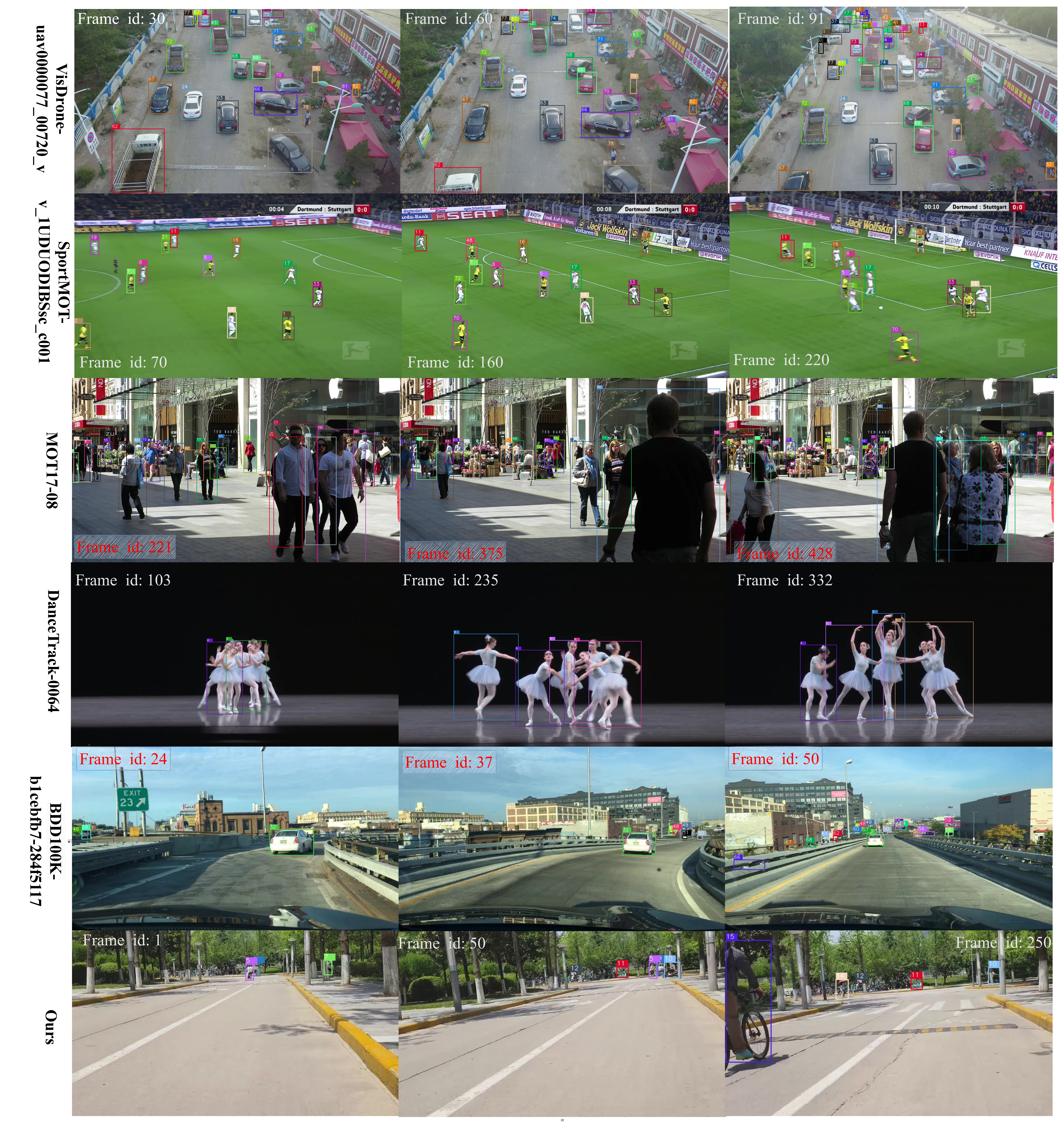}
	\end{center}
	\caption{The visualization of FastTrackTr on various datasets. Ours represents the actual measured data of our driverless car on the road, and the model used here is trained on BDD100k.}
	\label{visual_result}
\end{figure*}

\textbf{Comparison of State-Of-The-Art:} To better demonstrate the superiority of our FastTrackTr, we employed nearly all available data augmentations and used larger image sizes. Tables \ref{result_dancetrack}, \ref{result_MOT17} ,\ref{result_SportsMOT}, \ref{result_visdrone} and \ref{result_bdd100k} respectively showcase the maximum tracking performance of FastTrackTr on all benchmarks. Similar to many previous works \cite{zhou2020tracking,zhang2023motrv2,gao2023memotr,zeng2022motr}, we utilized the CrowdHuman \cite{shao2018crowdhuman} dataset for data augmentation on MOT17 and DanceTrack. It is important to note that although our method still does not outperform CNN-based models on the MOT17 benchmark, the underlying reason remains the well-discussed issue: for transformer models, the amount of data in the MOT17 dataset is too limited. Nevertheless, our method has surpassed all previous transformer-based approaches. This indicates that our method performs better even when the dataset images do not allow for sufficient training of the model, suggesting that our approach has potential for practical application in real-world settings, rather than being confined mostly to academic research like previous transformer methods, with few exceptions. And, We also provide various datasets and visualization results verified on physical objects, as shown in Fig. \ref{visual_result}.

\begin{table}
	\centering
	\caption{Performance comparison with state-of-the-art methods on MOT17 \cite{milanMOT16BenchmarkMultiObject2016}. The best performance among the transformer-based methods is marked in bold.}
	\label{result_MOT17}
\resizebox{\columnwidth}{!}{
\begin{tabular}{l|ccccc}
	\toprule
	\textbf{Methods} & \textbf{HOTA*} & \textbf{DetA} & \textbf{AssA} & \textbf{MOTA} & \textbf{IDF1} \\
	\rowcolor{gray!20}  \hline \textit{CNN based:} & & & & & \\
	FairMOT \cite{zhang2021fairmot} & 59.3 & 60.9 & 58.0 & 73.7 & 72.3 \\
    MMTrack\cite{xu2024rethinking} & 61.1 & - & - & 74.9 & 74.4 \\

	ByteTrack \cite{zhangByteTrackMultiobjectTracking2022} & 63.1 & 64.5 & 62.0 & 80.3 & 77.3 \\

	OC-SORT \cite{cao2023observation} & 63.2 & 63.2 & 63.4 & 78.0 & 77.5 \\
	DiffMOT \cite{lv2024diffmot} & 64.5 & 65.7 & 64.6 & 79.8 & 79.3 \\
	\rowcolor{gray!20}\hline \textit{Transformer based:} & & & & & \\
	TrackFormer \cite{meinhardt2022trackformer} & $/$ & $/$ & $/$ & 74.1 & 68.0 \\
	TransTrack \cite{sun2020transtrack} & 54.1 & 61.6 & 47.9 & 74.5 & 63.9 \\
	TransCenter \cite{xu2022transcenter} & 54.5 & 60.1 & 49.7 & 73.2 & 62.2 \\
	MeMOT \cite{cai2022memot} & 56.9 & $/$ & 55.2 & 72.5 & 69.0 \\
	MOTR \cite{zeng2022motr} & 57.2 & 58.9 & 55.8 & 71.9 & 68.4 \\
	MOTIP \cite{gao2024multiple} & 59.2 & 62.0 & 56.9 & 75.5 & 71.2 \\
	PuTR\cite{liu2024putr}  & 61.1 & $/$ & $/$ & 74.8 & 74.1 \\
	 
	 \rowcolor{green!8}\textbf{FastTrackTr(ours)}&\textbf{62.4}& \textbf{62.8}& \textbf{63.0} & \textbf{76.7} & \textbf{77.2} \\

	\hline
\end{tabular}
}

\end{table}

\begin{table}
	\centering
	\caption{Performance comparison with state-of-the-art methods on SportsMOT \cite{cui2023sportsmot}.}
	\label{result_SportsMOT}
	\resizebox{\columnwidth}{!}{
		\begin{tabular}{l|ccccc}
			\toprule
			\textbf{Methods} & \textbf{HOTA*} & \textbf{DetA} & \textbf{AssA} & \textbf{MOTA} & \textbf{IDF1} \\
			\hline FairMOT \cite{zhang2021fairmot} & 49.3 & 70.2 & 34.7 & 86.4 & 53.5 \\
			QDTrack \cite{fischer2023qdtrack} & 60.4 & 77.5 & 47.2 & 90.1 & 62.3 \\
			ByteTrack \cite{zhangByteTrackMultiobjectTracking2022} & 62.1 & 76.5 & 50.5 & 93.4 & 69.1 \\
			OC-SORT \cite{cao2023observation}  & 68.1 &  \textbf{84.8} & 54.8 & 93.4 & 68.0 \\
			MeMOTR \cite{gao2023memotr} & 68.8 & 82.0 & 57.8 & 90.2 & 69.9 \\
			\rowcolor{green!8}\textbf{FastTrackTr(ours)} & \textbf{70.1}  & 83.9  & \textbf{58.6} & \textbf{94.0} & \textbf{72.0} \\
			\hline
		\end{tabular}
	}
\end{table}

\begin{table}
\centering
\caption{Comparison With the SOTA Methods on the VisDrone2019 Test Set}
\label{result_visdrone}
\begin{tabular}{l|ccccc}
\hline \textbf{Methods}  & \textbf{MOTA} $\uparrow$ & \textbf{IDF1} $\uparrow$ & \textbf{FP} $\downarrow$ & \textbf{FN} $\downarrow$ & \textbf{IDS} $\downarrow$ \\
\hline MOTR \cite{zeng2022motr}  & 22.8 & 41.4 & 28407 & 147937 & 959 \\
TrackFormer \cite{meinhardt2022trackformer}  & 25.0 & 30.5 & 25856 & 141526 & 4840 \\
MMTrack \cite{xu2024rethinking}  & 36.7 & 54. 7 & 23849 & 120839 & 545 \\

\rowcolor{green!8} FastTrackTr  & \textbf{42.8} & \textbf{62.4} & \textbf{16474} & \textbf{85367} & \textbf{309} \\

\hline
\end{tabular}
\end{table}

\begin{table}
\centering
\caption{Comparison to existing methods on the BDD100K validation set. Best results are marked in bold.}
\label{result_bdd100k}
\begin{tabular}{ccccc}

\hline \textbf{Methods} & \textbf{TETA} & \textbf{LocA} & \textbf{AssocA} & \textbf{ClsA} \\
\hline 
QDTrack \cite{fischer2023qdtrack} & 47.8 & 45.8 & 48.5 & 49.2 \\
MOTRv2 \cite{zhang2023motrv2} & 54.9 & 49.5 & 51.9 & 63.1 \\
CO-MOT \cite{yanBridgingGapEndtoend} & 52.8 & 48.7 & \textbf{56 2} & 63.6 \\
\rowcolor{green!8} \textbf{FastTrackTr}  & \textbf{55.1} & \textbf{49.8} & 54 2 & \textbf{65.3}   \\

\hline
\end{tabular}
\end{table}

\subsection{Ablation Study}

We tested the impact of various components of our model and other factors on performance on the val set of \textbf{DanceTrack}.


The ablation study in Table \ref{ablation_all} highlights the progressive enhancements of FastTrackTr on the DanceTrack validation set. Starting with a baseline HOTA of 47.3, adding the Decoder increases HOTA to 51.6. Incorporating both Decoder and Encoder further boosts HOTA to 53.8. Finally, including the Mask module achieves the highest scores with HOTA of 54.1. In terms of the improvement degree, the Decoder with the introduction of historical information brings about the most significant performance enhancement. Adding an Encoder that further aggregates the temporal information of historical frames also contributes to a performance increase. The introduction of mask not only shields off useless historical information but also the randomness generated during its training can boost the generalization ability of the model.

\begin{table}
	\centering
	\caption{Ablation Study of our proposed FastTrackTr on the DanceTrack validation set.}
	\label{ablation_all}
	\resizebox{!}{!}{
		\begin{tabular}{c|c|c|c|c|c}\hline
			\textbf{Decoder}	& \textbf{Encoder} & \textbf{Mask} & \textbf{HOTA}  & \textbf{AssA}  & \textbf{IDF1} \\ \hline
			&  &  & 47.3 &  32.5 & 46.3 \\
			$\checkmark$ &  &&  51.6  &  37.2 & 51.3  \\
			$\checkmark$ & $\checkmark$ &   & 53.8  &  39.9 & 53.5  \\
			
			\rowcolor{green!8} $\checkmark$ & $\checkmark$ &  $\checkmark$ &  \textbf{54.1}  &  \textbf{40.0} & \textbf{54.7} \\
			\hline
		\end{tabular}
		
	}
\end{table}

In Table \ref{trajectory_attn}, we demonstrate the impact of different trajectory cross-attention mechanisms within the trajectory decoder on tracking performance. Here, $q^t$ represents the original query, while $q_f^{t-1}$ denotes the output of the decoder from the previous frame, with 'concat' indicating their concatenation. It is important to note that in this table, $q_f^{t-1}$ has not passed through the historical encoder, hence its performance is not as good as that of the final model. The advantage of Concat lies in combining the features of the current query and historical states. By leveraging the complementarity of spatiotemporal information and the dynamic balancing of the attention mechanism, it significantly boosts the joint performance of detection and tracking.

\begin{table}
	\caption {The Impact of Different Historical Cross Attention mechanisms on Tracking Performance }
	\label{trajectory_attn}
	\centering
	\resizebox{!}{!}{
		\begin{tabular}{cc|c|c|c|c}
			\hline $k$ & $v$  & \textbf{HOTA} & \textbf { DetA }  & \textbf { AssA }  & \textbf { IDF1 }  \\
			\hline
			$q^t$ & $q_f^{t-1}$ & 48.8 & 70.4 & 35.8 & 48.1   \\
			$q_f^{t-1}$ & $q_f^{t-1}$ & 48.2 & 70.0 & 33.0 & 47.7   \\
			$q_f^{t-1}$ & $q^t$ & 49.0 & 71.5 & 35.9 & 48.5   \\
			\rowcolor{green!8} concat & concat &  \textbf{51.6}  &\textbf{72.8}&  \textbf{37.2} & \textbf{51.3} \\
			\hline
		\end{tabular}
	}
\end{table}	



In Table \ref{loss_ablation}, we demonstrate the impact of using circle loss \cite{sun2020circle}, triplet loss \cite{schroff2015facenet}, and one-hot encoding loss of the JDE \cite{wang2020towards} on our method performance during the training of ID embeddings. From the result, circle loss consistently outperforms both triplet loss and one-hot encoding loss across all metrics. Specifically, circle loss achieves the highest tracking performance. In comparison, triplet loss shows the lowest performance, and one-hot encoding loss provides intermediate results. These results demonstrate that circle loss effectively enhances the training of ID embeddings for, leading to superior tracking performance in our method.

\begin{table}
	\caption {The impact of various ID Embeddings losses on performance.}
	\centering
	\label{loss_ablation}
		\begin{tabular}{c|c|c|c|c}
			\hline \textbf { $loss$ } & \textbf { HOTA } & \textbf { DetA }  & \textbf { AssA }  & \textbf { IDF1 }  \\
			\hline

			triplet loss & 52.7  & \textbf{76.8}& 38.1 & 52.4  \\
			one hot & 53.8  & 75.2& 39.4 & 54.1  \\
			\rowcolor{green!8}circle loss & \textbf{54.1}  & 75.6& \textbf{40.0} & \textbf{54.7}  \\

			\hline
		\end{tabular}
\end{table}

\subsection{Experiments in Actual Application Scenarios}

\begin{figure}[t]
	\begin{center}
		\includegraphics[width=0.6\linewidth]{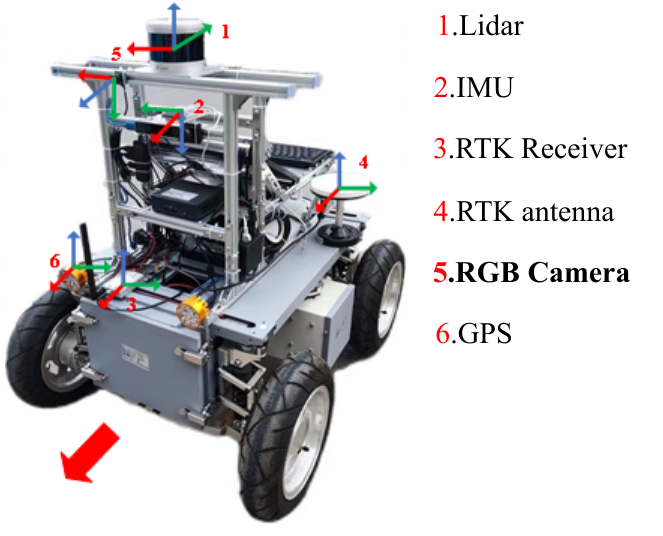}
	\end{center}
	\caption{Unmanned vehicle for physical verification.}
	\label{auto_car}
\end{figure}

\begin{table}
	\centering
	\caption{The speed and accuracy of our method and some existing methods using TensorRT FP16 precision on edge devices on our own verification platform and the accuracy of these models on Dancetrack are provided for reference. }
    \label{latency}

	\resizebox{1\linewidth}{!}{
\begin{tabular}{l|c|c|c}
	\hline \textbf{Methods} & \textbf{Image Res.} & \textbf{Overall Latency(ms)} & \textbf{HOTA} \\
	\hline \text {Deep OC-SORT}\cite{maggiolinoDeepOCSORTMultiPedestrian2023} & 1333 $\times$ 800 & 163.9 ms(6.1 fps)&  58.5\\
	\text {PuTR }\cite{liu2024putr} & 1333 $\times$ 800 & 67.3ms(14.8fps)&  54.5\\
	 \text {OC-SORT} \cite{cao2023observation} & 1333 $\times$ 800 & 54.3 ms(18.4 fps)& 52.1 \\
	 
	 \rowcolor{green!8} \text {FastTrackTr} & 1333 $\times$ 800 & 48.7 ms(20.5 fps)&  56.9\\
	 
	 \text {Deep OC-SORT} \cite{maggiolinoDeepOCSORTMultiPedestrian2023} & 640 $\times$ 640 & 141.1 ms(7.1 fps)& - \\
		\text {OC-SORT} \cite{cao2023observation} & 640 $\times$ 640 & 36.7 ms(27.2 fps)& - \\
	 \rowcolor{green!12} \text {FastTrackTr} & 640 $\times$ 640 & 35.6 ms(28.1 fps)&  54.8\\
	
	 \rowcolor{green!16} FastTrackTr-Dec3 & 640 $\times$ 640 & 30.1ms(33.2fps)&  51.2\\
	 \rowcolor{green!22}  FastTrackTr-R18-Dec3 & 640 $\times$ 640  & 16.8ms(59.2fps)& 47.4 \\
	\hline
\end{tabular}
}
\end{table}

To validate the performance of our method in practical industrial environments, we conducted tests on the autonomous vehicle shown in Fig. \ref{auto_car}. The core processor of this autonomous vehicle is NVIDIA's Jetson AGX Orin, and the vehicle is equipped with sensors such as a lidar, an IMU, a GPS module, a fisheye camera, and an RGB camera. A Redmi AX3000 router is used to build an in-vehicle local area network, satisfying the network communication between the host and the lidar as well as the data transmission between the client and the autonomous vehicle host. To further simulate real-world scenarios, in addition to the tracking method, a SLAM algorithm is also running on the autonomous vehicle. The final time consumption of each tracking method is shown in Table \ref{latency}, and the indicators marked in the table are the accuracy of the methods on Dancetrack. Since the official details on how OC - SORT trains the YOLOX detector are not provided, the accuracy of our trained model is not ideal, which may cause some misunderstandings. Therefore, we do not provide the specific tracking accuracy of OC - SORT and Deep OC - SORT under the image size of 640×640, and only provide the speed.

\section{Conclusion}
In this paper, we introduce a real time tracking method, FastTrackTr, which demonstrates superior performance across multiple datasets. By incorporating novel trajectory encoder and decoder designs, we effectively enhance the accuracy and robustness of tracking. Additionally, FastTrackTr achieves significant improvements in computational efficiency, presenting substantial potential for real-time applications. Future work will focus on further optimizing the model structure to adapt to more complex scenarios and exploring deployment possibilities across different hardware platforms. However, our method currently does not surpass the tracking accuracy of MOTIP, primarily due to its specialized network architecture for ID encoding and sequence-optimized training paradigm that captures long-term temporal dependencies. which ensures better temporal information learning. As the saying goes, "there are trade-offs"; these aspects also result in poor inference speed and slow training for MOTIP. Conversely, FastTrackTr does not exhibit these drawbacks.


\bibliographystyle{IEEEtran}
\bibliography{IEEEabrv,natbib}

\end{document}